\documentclass{article}

\usepackage{mathptmx}
\usepackage{soul}\setuldepth{article}

\def\hb{\hbox to 11.5 cm{}}

\usepackage{todonotes}
\usepackage{booktabs}
\usepackage{amsthm,amsmath}
\usepackage{subfig}
\usepackage{url}

\newcommand{\afgcn}{\texttt{AFGCN}}
\newcommand{\afgat}{\texttt{AFGAT}}
\newcommand{\harper}{\texttt{HARPER++}}
\newcommand{\aripoter}{\texttt{ARIPOTER}}
\newcommand{\fargo}{\texttt{FARGO-LIMITED}}

\theoremstyle{definition}
\newtheorem{definition}{Definition}
\newtheorem{example}{Example}

\newcommand{\af}{\mathcal{F}}
\newcommand{\args}{\mathcal{A}}
\newcommand{\atts}{\mathcal{R}}

\newcommand{\semantics}[1]{\mathbf{#1}}
\newcommand{\complete}{\semantics{co}}
\newcommand{\preferred}{\semantics{pr}}
\newcommand{\stable}{\semantics{st}}
\newcommand{\grounded}{\semantics{gr}}
\newcommand{\hcat}{\semantics{h-cat}}
\newcommand{\Mbs}{\semantics{Mbs}}
\newcommand{\Cbs}{\semantics{Cbs}}
\newcommand{\nsa}{\semantics{nsa}}
\newcommand{\cred}{\semantics{cred}}
\newcommand{\skep}{\semantics{skep}}

\newcommand{\complexity}[1]{\textsc{#1}}
\renewcommand{\P}{\complexity{P}}
\newcommand{\NP}{\complexity{NP}}
\newcommand{\coNP}{\complexity{coNP}}

\newcommand{\problem}[1]{\textbf{#1}}
\newcommand{\DC}{\problem{DC}}
\newcommand{\DS}{\problem{DS}}

\title{Graph Convolutional Networks and Graph Attention Networks for Approximating Arguments Acceptability -- Technical Report}
\author{Paul Cibier\\Independent researcher, Paris, France\\paul.cibier@gmail.com \and Jean-Guy Mailly\\IRIT, Universit\'e Toulouse Capitole, Toulouse, France\\ jean-guy.mailly@irit.fr}

\begin{document}

\maketitle

\begin{abstract}
Various approaches have been proposed for providing efficient computational approaches for abstract argumentation. Among them, neural networks have permitted to solve various decision problems, notably related to arguments (credulous or skeptical) acceptability. In this work, we push further this study in various ways. First, relying on the state-of-the-art approach \afgcn, we show how we can improve the performances of the Graph Convolutional Networks (GCNs) regarding both runtime and accuracy. Then, we show that it is possible to improve even more the efficiency of the approach by modifying the architecture of the network, using Graph Attention Networks (GATs) instead.
\end{abstract}


Computational methods for abstract argumentation \cite{Dung95} have received much attention in the last years, notably thanks to the organization of the International Competition on Computational Models of Argumentation (ICCMA).\footnote{\url{http://argumentationcompetition.org}} Roughly speaking, there are two families of approaches: exact algorithms, which guarantee to find the correct result, but possibly require a large amount of time (because of the high computational complexity of many interesting problems \cite{HOFAComplexity}); and approximate algorithms, which typically outperform exact algorithms in terms of runtime, but may provide wrong results in some cases.

Several recent approaches have been proposed for defining approximate algorithms for abstract argumentation. Some of them rely on the relation between the grounded semantics and most other classical semantics \cite{CeruttiTV20}, which allows to use this (polynomially computable) semantics as a good approximation for the status of most arguments under other (intractable) semantics. In the recent editions of ICCMA, these approaches are \harper{} \cite{Thimm2023c} and \aripoter{} \cite{DelobelleMR23}. Another approach, named \fargo{} \cite{Thimm2023b}, is based on (bounded depth) DPLL-style algorithm %\cite{DavisLL62} 
for searching if an argument belongs to an admissible set, which allows to answer positively to various acceptability queries.
Finally, using machine learning techniques for abstract argumentation has received some attention, in particular \afgcn{} \cite{MalmqvistYNM20} is based on Graph Convolutional Networks (GCNs). All these approaches participated to the last edition of ICCMA in 2023,\footnote{\url{http://argumentationcompetition.org/2023}} and solve credulous and skeptical acceptability problems, {\em i.e.} determining whether a given argument belongs to some or each extension for a given semantics.

In this work, we explore two different ways to define approximate algorithms for these acceptability queries, both based on machine learning techniques. First, building upon \afgcn, we define new GCN-based methods for solving acceptability queries. We show that these methods, using different sets of features for training the model or different network architectures allow to outperform the original \afgcn{} solver. Then, we show how replacing GCNs by Graph Attention Networks (GATs) improves even more the performances of the solver.

\section{Background}
\subsection{Abstract Argumentation}

\begin{definition}[Argumentation Framework \cite{Dung95}]
An {\em Argumentation Framework} (AF) is a directed graph $\af = \langle \args, \atts\rangle$ where $\args$ is a set of abstract entities called {\em arguments} and $\atts \subseteq \args \times \args$ is the {\em attack relation}.
\end{definition}

Although Dung does not put additional constraints regarding the set of arguments, in this work we assume that $\args$ is a non-empty finite set of arguments. When $(a,b) \in \atts$, we say that $a$ {\em attacks} $b$, and similarly if $\exists a \in S$ s.t. $a$ attacks $b$, then the set of arguments $S$ attacks $b$. Classical AF semantics rely on a notion of collective acceptability: the semantics allow to determine sets of {\em extensions}, which are sets of jointly acceptable arguments. Most extension-based semantics satisfy two basic properties:

\begin{definition}[Conflict-freeness and defense]
Given an AF $\af = \langle \args, \atts \rangle$ and the set of arguments $S \subseteq \args$, we say that $S$ is {\em conflict-free} if $\forall a, b \in S$, $(a,b) \not \in \atts$. Then, given  an argument $a \in \args$, we say that $S$ {\em defends} $a$ if $\forall b \in \args$ s.t. $(b,a) \in \atts$, $\exists c \in S$ s.t. $(c,b) \in \atts$.
\end{definition}

A conflict-free set which defends all its elements is said {\em admissible}. Most semantics select extensions among the admissible sets of an AF.

\begin{definition}[Dung's Semantics \cite{Dung95}]
Given $\af =\langle \args, \atts \rangle$ and $S \subseteq \args$ an admissible set, 
\begin{itemize}
    \item $S$ is a {\em complete} extension iff it contains all the arguments that it defends,
    \item $S$ is a {\em preferred} extension iff it is a $\subseteq$-maximal complete extension,
    \item $S$ is a {\em grounded} extension iff it is a $\subseteq$-minimal complete extension,
    \item $S$ is a {\em stable} extension iff it attacks every argument in $\args \setminus S$.
\end{itemize}
\end{definition}

We write (resp.) $\complete(\af)$, $\preferred(\af)$, $\grounded(\af)$ and $\stable(\af)$ for the sets of complete, preferred, grounded and stable extensions of $\af$. It is known \cite{Dung95} that each AF has exactly one grounded extension, at least one preferred and complete extension, and $\stable(\af) \subseteq \preferred(\af) \subseteq \complete(\af)$. The status of an argument w.r.t. a given semantics is defined by:

\begin{definition}[Argument Acceptability]
Given an AF $\af = \langle \args, \atts\rangle$ and a semantics $\sigma \in \{\complete,\preferred,\grounded,\stable\}$, the argument $a \in \args$ is {\em credulously} (resp. {\em skeptically}) accepted if $a \in \cred_\sigma(\af) =  \bigcup_{S \in \sigma(\af)} S$ (resp. $a \in \skep_\sigma(\af) = \bigcap_{S \in \sigma(\af)} S$).
\end{definition}

Reasoning with the grounded semantics is tractable, but it is not the case in general for the other semantics. More precisely, $\DC$-$\sigma$ (resp. $\DS$-$\sigma$) is the decision problem which consists in determining whether an argument $a$ is credulously (resp. skeptically) accepted w.r.t. the semantics $\sigma$. $\DC$-$\sigma$ is \NP-complete for $\sigma \in \{\complete, \preferred, \stable\}$, and $\DS$-$\sigma$ is \coNP-complete for $\sigma = \stable$ and $\Pi_2^\P$-complete for $\sigma = \preferred$ (it is polynomial for $\sigma = \complete$ because it coincides with $\DS$-$\grounded$). See \cite{HOFAComplexity} for more details on this topic.

\begin{example}
Figure~\ref{fig:example-af} depicts an AF $\af_1 = \langle \args_1, \atts_1 \rangle$. Its extensions and the set of (credulously and skeptically) accepted arguments are given in Figure~\ref{tab:example-extensions}.
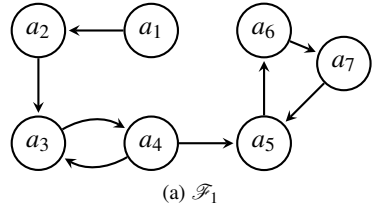
\begin{figure}[htb]
  \centering
  \subfloat[$\af_1$\label{fig:example-af}]{
  \scalebox{1}{
  \begin{tikzpicture}[->,>=stealth,shorten >=1pt,auto,node distance=1.5cm,
                thick,main node/.style={circle,draw,font=\bfseries}]
\node[main node] (a1) {$a_1$};
\node[main node] (a2) [left of=a1] {$a_2$};
\node[main node] (a3) [below of=a2] {$a_3$};
\node[main node] (a4) [right of=a3] {$a_4$};
\node[main node] (a5) [right of=a4] {$a_5$};
\node[main node] (a6) [above of=a5] {$a_6$};
\node[main node] (a7) [above right of=a5] {$a_7$};

\path[->] (a1) edge (a2) 
    (a2) edge (a3)
    (a3) edge[bend left] (a4)
    (a4) edge[bend left] (a3)
    (a4) edge (a5)
    (a5) edge (a6)
    (a6) edge (a7)
    (a7) edge (a5);
  \end{tikzpicture}
} % End scalebox
} % End subfloat

\subfloat[Extensions and (credulously and skeptically) acceptable arguments \label{tab:example-extensions}]{
\resizebox{0.7\linewidth}{!}{
 \begin{tabular}{cccc}
    \toprule
        Semantics $\sigma$ & $\sigma(\af_1)$ & $\cred_\sigma(\af_1)$ & $\skep_\sigma(\af_1)$ \\
        \midrule
        $\complete$ & $\{\{a_1, a_4, a_6\}, \{a_1, a_3\}, \{a_1\}\}$ & $\{a_1, a_3, a_4, a_6\}$ & $\{a_1\}$ \\
        $\preferred$ & $\{\{a_1, a_4, a_6\}, \{a_1, a_3\}\}$ & $\{a_1, a_3, a_4, a_6\}$ & $\{a_1\}$ \\
        $\grounded$ & $\{\{a_1\}\}$ & $\{a_1\}$ & $\{a_1\}$ \\
        $\stable$ & $\{\{a_1, a_4, a_6\}\}$ & $\{a_1, a_4, a_6\}$ & $\{a_1, a_4, a_6\} $\\
    \bottomrule
    \end{tabular}
} % End resizebox
} % End subfloat
  \caption{An AF $\af_1$ with its extensions and accepted arguments under  $\sigma \in \{\complete, \preferred, \grounded, \stable\}$. }
\end{figure}

\end{example}

While extension-based semantics rely on a notion of collective acceptability of arguments (and then, individual acceptability can be derived thanks to credulous or skeptical reasoning), other families of semantics directly capture a notion of individual acceptability. In particular, a {\em gradual semantics} assigns to each argument in an AF an {\em acceptability degree}, usually a real number in the interval $[0,1]$, where $0$ represents complete rejection of the argument and $1$ complete acceptance. These semantics are based on different intuitions, for instance taking into account the number or the quality of an argument's attackers. To represent the attackers of an argument $a$, we define $a^- = \{b \in \args \mid (a,b) \in \atts\}$.
Here are some classical gradual semantics that we use in this work.

\begin{definition}[Gradual Semantics]
Given an AF $\af = \langle \args, \atts \rangle$ and $a \in \args$, the gradual semantics {\em h-categorizer} ($\hcat$ \cite{BesnardH01}), {\em no self-attacker} ($\nsa$ \cite{BeuselinckDV23}), {\em Max-based} ($\Mbs$ \cite{AmgoudBDV17}) and {\em Card-based} ($\Cbs$ \cite{AmgoudBDV17}) map each argument $a \in \args$ to a value in $[0,1]$ as follows:
\[
\hcat(\af,a) = \frac{1}{1 + \sum_{b \in a^-} \hcat(\af,b)}
\]

\[
\nsa(\af,a) = 
\left \{
\begin{array}{rl}
    0 & \text{if } (a,a) \in \atts\\
    \frac{1}{1 + \sum_{b \in a^-} \nsa(\af,b)} & \text{otherwise}
\end{array}
\right .
\]

\[
\Mbs(\af, a) = \frac{1}{1 + \max_{b \in a^-} \Mbs(b,\af)}
\]

\[
\Cbs(\af, a) = \frac{1}{1 + |a^-| + \frac{\sum_{b \in a^-} \Cbs(\af,b)}{|a^-|}}
\]
\end{definition}

\begin{example}\label{example:gradual-semantics}
Let $\af_2 = \langle \args_2, \atts_2\rangle$ be the AF from Figure~\ref{fig:example-af-gradual}. For each $x \in \args_2$, we give the acceptability degree for all the gradual semantics considered in the paper in Table~\ref{tab:example-gradual-semantics}.

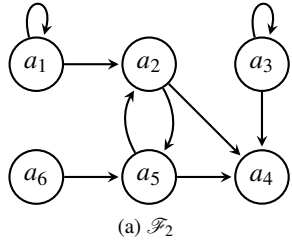
\begin{figure}[htb]
  \centering
  \subfloat[$\af_2$ \label{fig:example-af-gradual}]{
\scalebox{1}{
  \begin{tikzpicture}[->,>=stealth,shorten >=1pt,auto,node distance=1.5cm,
                thick,main node/.style={circle,draw,font=\bfseries}]
\node[main node] (a1) {$a_1$};
\node[main node] (a2) [right of=a1] {$a_2$};
\node[main node] (a3) [right of=a2] {$a_3$};
\node[main node] (a4) [below of=a3] {$a_4$};
\node[main node] (a5) [left of=a4] {$a_5$};
\node[main node] (a6) [left of=a5] {$a_6$};

\path[->] (a1) edge[loop above] (a1) 
    (a1) edge (a2) 
    (a2) edge[bend left] (a5)
    (a2) edge (a4)
    (a3) edge[loop above] (a3)
    (a3) edge (a4)
    (a5) edge[bend left] (a2)
    (a5) edge (a4)
    (a6) edge (a5);
  \end{tikzpicture}
} % End scalebox
  } % End subfloat
  
  \subfloat[Acceptability degrees (rounded to $10^{-3}$) \label{tab:example-gradual-semantics}]{
  \resizebox{0.6\linewidth}{!}{
    $
    \begin{array}{c  c  c  c  c}
    \toprule
    \text{Argument } x\text{ } & \text{ }\hcat(\af_2,x)\text{ } & \text{ }\nsa(\af_2,x)\text{ } & \text{ }\Mbs(\af_2,x)\text{ } & \text{ }\Cbs(\af_2,x) \\
    \midrule
    a_1 & 0.618 & 0.0 & 0.618 & 0.414 \\
    a_2 & 0.495 & 0.732 & 0.618 & 0.299 \\
    a_3 & 0.618 & 0.414 & 0.618 & 0.414 \\
    a_4 & 0.398 & 0.477 & 0.618 & 0.231 \\
    a_5 & 0.401 & 0.399 & 0.5 & 0.274 \\
    a_6 & 1.0 & 1.0 & 1.0 & 1.0 \\
    \bottomrule
    \end{array}
    $
    } % End resizebox
  } % End subfloat
  \caption{An AF $\af_2$ and the acceptability degrees of arguments for $\sigma \in \{\hcat, \nsa, \Mbs, \Cbs\}$.}
\end{figure}
\end{example}

\subsection{Neural Networks for Argumentation}

Now we present basic notions of deep learning as used in the rest of this paper. Deep learning is an efficient way to approximate the solutions to various kinds of problems \cite{DongWA2021}. Given the high complexity of most interesting problems in abstract argumentation \cite{HOFAComplexity}, it is not surprising that several approximate reasoning approaches have been proposed for argumentation, including some based on deep learning.

An artificial neuron (or simply, a neuron from now) can be seen as a computational unit taking as input a vector of values $x = (x_1, \dots, x_n)$, a vector of weights $w = (w_1,\dots,w_n)$ and an additional information called the bias $b$. The neuron is mainly an {\em activation function} $f$, and the output of the neuron is obtained by applying $f$ to the weighted sum of the input values and the bias, {\em i.e.} $f(b + \sum_{i = 1}^n w_i \times v_i)$. Neurons are connected into networks, usually made of various {\em layers}. The {\em input layer} is made of neurons that only pass their input values to the next layer without processing ({\em i.e.} there is no activation function), then there are (possibly many) hidden layers and finally one output layer, s.t. the output of these last neurons correspond to the output of the neural network. For training a neural network in a context of supervised learning, the output of the network is compared to the labels of the training data, and a back-propagation algorithm is used to find the most accurate representation of the data by adjusting the weights of the neurons. This back-propagation aims at minimizing the value of the loss function, which represents the distance between the value predicted by the network output and the real value of the labeled data. This allows to provide a better approximation of the mapping (input data $\rightarrow$ labels) by the neural network. Neural networks are typically represented as (weighted) directed graphs where the nodes are the neurons and the weighted edges are the connections between neurons.

\subsubsection{Graph Convolutional Network}

While ``basic'' neural networks as described before take a vector of  real numbers as input, \cite{KipfW17} introduces neural networks able to directly use graphs as their inputs, named Graph Convolutional Networks (GCNs). A GCN works by taking as input an adjacency matrix representation of the graph and a node embedding, {\em i.e.} a set of features for each node of the graph.
A layer of a GCN works by applying a convolution on each dimension of the vector representing the node embedding. The convolution takes into account the node embedding of the neighbours of the node, and the node itself. Formally, from a layer $l$ of the GCN and the adjacency matrix of the graph $A$, the layer $l+1$ can be obtained by computing $H^{l+1} = f(H^l,A) = \sigma(\hat{D}^{-\frac{1}{2}} \hat{A} \hat{D}^{-\frac{1}{2}} H^l W^l)$ where $\hat{A}$ is the representation of the graph where self-loops have been added on each argument,
$W^l$ is the matrix of weights for the layer $l$ and $\sigma$ is a (non-linear) activation function. The self-loops  allow a node of the argumentation framework to propagate information to itself (otherwise, the node would only receive information about its neighbours). The first layer $H^1$ can be obtained by computing $f(H^0,A)$ where $H^0$ is a matrix representation of the node features.

The GCN model was used in the context of abstract argumentation for approximating the credulous or skeptical acceptability of arguments under various semantics. \cite{KuhlmannT19} focused on credulous acceptability under the preferred semantics ($\DC$-$\preferred$) and proposed two different node embeddings. In the first one, each node is associated to a single feature, which is the same constant for every node ({\em i.e.} there is no additional information provided for the nodes), while the second one provides two features by node, namely the numbers of incoming and outgoing attacks. Then, another approach based on GCNs has been proposed, with the solver \afgcn{} performing extremely well at ICCMA 2021.\footnote{\url{http://argumentationcompetition.org/2021/}} The second version of \afgcn, which participated to ICCMA 2023,\footnote{\url{http://argumentationcompetition.org/2023/}} can solve both $\DC$ and $\DS$ problems for the semantics $\complete$, $\preferred$ and $\stable$, as well as other semantics not considered in this paper (semi-stable, stage and ideal \cite{HOFASemantics}). We give more details on \afgcn{} in Section~\ref{section:contributions}, where we describe the various modifications that we have made in order to improve its efficiency regarding both runtime and precision.

\subsubsection{Graph Attention Network}
In classical graph neural networks like GCNs, the feature update of a node when passing from layer $l$ to layer $l+1$ is typically an average of the features of all its neighbors and itself. It means that there is no difference in the treatment of all neighbors. With Graph Attention Networks (GATs) \cite{VelickovicCCRLB18}, there is an assumption that some node are more important than some other ones. To take into account this assumption, when updating the node embedding of the node in a graph attentional layer, an attention score is computed between the source node and every neighbour of this node, which allows to indicate if a node $j$ is important or not for a node $i$. These attention coefficients are used to weigh the influence of a neighbors' features when computing the new embedding of the node during the convolution.
The authors of \cite{VelickovicCCRLB18} also incorporate the mechanism of multi-head attention, which allows the network to perform the same process described above independently and in parallel. When finishing the computation of the updated features with each attention head, the results of all the heads are concatenated (for inner layers) or averaged (for output layers).

\section{New Neural Networks for Arguments Acceptability}\label{section:contributions}
In this paper we propose several new approaches to (approximatively) resolve arguments acceptability task for abstract argumentation. Section~\ref{section:contributions-gcn} shows how we have modified the state-of-the-art \afgcn{} approach for improving its performances, and Section~\ref{section:contributions-gat} shows our proposal for using GATs instead of GCNs.

\subsection{GCNs for Arguments Acceptability}\label{section:contributions-gcn}

Our first approach is based on the work by Lars Malmqvist using Graph Convolutional Networks \cite{MalmqvistYNM20} for approximating arguments acceptability, but (among other technical differences) our method uses a different kind of node embedding. Based on \cite{MalmqvistYNM20} and the solver source code\footnote{\url{https://github.com/lmlearning}}, let us first describe with more details the \afgcn{} solver, more precisely its second version that was submitted to ICCMA 2023.
\afgcn v2 starts by running a grounded extension solver (based on NumPy \cite{harris2020array}, which provides a quadratic space representation of the AF), and if the query argument is not a member of the grounded extension, then a GCN is used to predict the acceptability of this argument. The GCN model is built as follow. Its inputs are the adjacency matrix of the argumentation graph and a graph embedding build made of a vector of dimension $128$ for each argument, containing features representing the eigenvector centrality, the graph centrality, the in-degree and out-degree, PageRank and the graph coloring. The other features are randomized input features.
All of these assign numbers to the nodes, which are normalized in order to belong to a comparable scale and constitute the node embedding of the graph.

The core component of the GCN is built with $4$ consecutive blocks, each made of a GCN layer and a Dropout layer \cite{SrivastavaHKSS14}, which aim at reducing the risk of overfitting by dropping some of the neurons during the forward pass of the learning process. At each block, there are residual connections that feed it with the original features (in addition to the new embedding produced by the previous layer).
After these blocks, a layer reduces the dimension of the features vector to $1$.
Finally, a Sigmoid layer computes the probability for the arguments to be accepted.

\subsubsection{Speed and Memory Management Improvements}\label{section:technical-improvements}
There are several parts of the process where \afgcn v2 could not succeed its computation because of output errors or timeout, notably when managing large instances.

Python provides various interesting tools for applying machine learning techniques, but it may lack of efficiency in some cases. In particular, for \afgcn v2, the time taken for computing the graph centrality when building the node embedding may be important, as well as the time required to parse the text files describing large instances.

To improve the speed of the solver, we propose a Rust Made Python package that can read and parse the AF much faster than the Python implementation. This package receives a path to the file describing the AF, and then parses it. The second part of the process is the computation of its grounded extension. This Rust tool verifies if the query argument belongs to the grounded extension or is attacked by it. In this case, the system immediately stops and provides the correct answer.
If the argument is neither in the grounded extension nor attacked by it ({\em i.e.} it is labeled UNDEC regarding the grounded labelling of the AF \cite{Caminada06}), then we need to use the neural network to estimate the acceptability of the arguments. So, the Rust package computes the node embedding of the argumentation graph in order to feed the GCN or kind of GNN that need a node embedding.

This approach can speed up a lot the process for several reasons. As said before, the Rust package is faster for parsing large AF files compared to the original Python implementation. The main improvement regarding runtime is computation of the graph centrality, which was the main cause for long runtimes (and timeout) with \afgcn v2. Its computation is much faster with the Rust package than with the Python-based NetworkX library \cite{HagbergSS08} used by \afgcn v2. The other potential runtime improvement concerns arguments attacked by the grounded extension. These arguments cannot belong to any extension under any semantics considered in this paper (nor other classical semantics like the semi-stable and ideal semantics \cite{HOFASemantics}), but \afgcn v2 still needs computing the node embedding and running the GCN for them, while our package instantly answers that these arguments are not acceptable. This approach, inspired by \cite{Thimm2023c,DelobelleMR23} can considerably speed up the computation for arguments attacked by the grounded extension.

Another issue with the computation of the grounded extension is the use of the quadratic data structure in \afgcn v2. This leads to some {\em out-of-memory} errors with large instances (for instance, an AdmBuster graph\footnote{See \url{https://argumentationcompetition.org/2017/AdmBuster.pdf} for details on AdmBuster.} with $n=50000$, requires 18.6 GiB of memory). On the contrary, our Rust package relies on a linear algorithm and data structure, similar to the approach from \cite{NofalAD21}, which avoids errors like those faced by \afgcn v2.

\subsubsection{Model Precision Improvement}
The node embedding of a graph consist in mapping each node (in our case, each argument) of the graph with a vector of $N$ dimensions, each dimension representing information about some features of the node. As explained previously, in \afgcn, $N = 128$ with a large part of the features being randomly generated. We propose several variants of the node embedding of \afgcn, where the acceptability degrees of the argument for various gradual semantics (namely, $\hcat$, $\nsa$, $\Cbs$ and $\Mbs$) are added to the features of the argument. Another feature that we take into account is the acceptability status w.r.t. the grounded semantics (represented as $1$ for arguments in the grounded extension, $0$ for arguments attacked by it, and $0.5$ for the remaining arguments).

In the following, we describe five versions of \afgcn:
\begin{itemize}
    \item \afgcn v2, {\em i.e.} the solver that participated to ICCMA 2023, (with the features corresponding to the eigenvector centrality, the graph centrality, the in-degree and out-degree, PageRank, the graph coloring, and the randomised values),
    \item \afgcn-P128, our implementation (with the technical improvements described in Section~\ref{section:technical-improvements}) with the $6$ features used in \afgcn v2 and the $5$ additional features corresponding to the gradual semantics and the grounded semantics, and $117$ randomly generated features,
    \item \afgcn-P11, our implementation with only the $11$ ``meaningful'' features ({\em i.e.} without the random features).
    \item \afgcn-P128$^{-do}$, similar to \afgcn-P128 but without dropout layers,
    \item \afgcn-P11$^{-do}$, similar to \afgcn-P11 but without dropout layers.
\end{itemize}

Comparing \afgcn v2 with \afgcn-P128 aims at evaluating the interest of the new features ({\em i.e.} the gradual semantics and the grounded semantics), while the comparison between \afgcn-P128 and \afgcn-P11 assesses the impact of the randomized features. These randomized features are used for preventing the dropout layers from having a negative impact on the learning process by dropping too many important features. For this reason, we also perform a comparison of these approaches without the dropout layers.

\subsection{GATs for Arguments Acceptability}\label{section:contributions-gat}
We have also implemented a Graph Attention Network (called \afgat{} in the rest of the paper), and used the second version proposed by the authors \cite{VelickovicCCRLB18}, which generally outperforms the first version in all benchmarks according to the authors of the model.
Intuitively, not all neighbours of an arguments have the same impact on its acceptability ({\em e.g.} an attacker of $a$ which is attacked by the grounded extension has no impact on $a$, but an attacker which is UNDEC in the grounded labelling may belong to some extension and thus prevent $a$ from being skeptically accepted), hence the interest of the attention mechanism of GATs for evaluating arguments acceptability.
As far as we know, it is the first implementation of a GAT for argument acceptability according to extension-based semantics. 
In our experiment we choose to implement the GATv2 with 3 graph attentionals layers parameterized with 5 head for the first layer, and 3 heads for the second and the third ones. We only consider the same 11 input features as in \afgcn-P11. For the following layers, the numbers of their input features is the number of output features from the previous layer multiplied by the number of attention heads from the previous layer (so there are $5 \times 5 = 25$ features for the second layer, and $5 \times 3 = 15$ for the third layer), while the output layer only has one feature representing the acceptability of the arguments we have an Sigmoid Layer to transform the output feature of the last layer into a probability of acceptability.
Following the approach described by \cite{VelickovicCCRLB18}, the concatenation of the results provided by each attention head is used between the layers 1 and 2 on the one hand, and 2 and 3 on the other hand.
The average operation is used after the last layer because according to the directive of the authors the use average operation on the last layer when we use multi-head attention on it.
\section{Experimental Evaluation}
In the following, we compare the efficiency of our approaches with those of \harper{} and \fargo{} which were the best performers in the approximate track of ICCMA 2023, as well as \afgcn{} which is the state-of-the-art approximate solver based on neural networks (and the foundation of our work).

\subsection{Protocol}

All our GNNs ({\em i.e.} \afgcn-P11, \afgcn-P128, \afgcn-P11$^{-do}$, \afgcn-P128$^{-do}$ and \afgat) were trained with the same dataset used to train \afgcn v2 by Lars Malmqvist, {\em i.e.} an aggregate of instances from ICCMA 2017, 900 AFs from the sets A, B, C\footnote{\url{https://github.com/lmlearning/AFGraphLib}}. %all 5 difficulty categories.
Similarly to \afgcn v2, we used the Deep Graph Library (DGL) and Pytorch to train our models, with the option ``shuffle'' enabled to allow DGL to reshuffle the set of graphs given to the GNN for every batch during training. The batch size was set to $64$ for all variants of \afgcn{} and $4$ for \afgat{} (because of memory limitations).
For the optimization process, we used Adam \cite{KingmaB14} (like in \afgcn v2). The learning rate was set to $0.01$, like \afgcn v2. 
The training ended after $400$ epochs.

The hardware used for the training was a CPU i3-9100F, 8GB of RAM and a GPU Nvidia GTX 1070 (8GB VRAM). For the test we used the same configuration except that solvers did not have access to the GPU for inference ({\em i.e.} it was only run on the CPU).

\subsection{Results}

Now, we describe our experimental results regarding four classical decision problems, namely $\DC$-$\complete$, $\DC$-$\stable$, $\DS$-$\preferred$ and $\DS$-$\stable$. We conducted two different evaluations. In both cases, we used the instances from the ICCMA 2023 competition (329 instances).\footnote{\url{https://zenodo.org/records/8348039/files/iccma2023_benchmarks.zip}}

\subsubsection{Comparison of the Neural Networks}
In this first experiment, in order to define our test set we tried to compute with the SAT-based solver Crustabri\footnote{\url{https://github.com/crillab/crustabri}} the acceptability status for each argument in each AF in the dataset. Notice that in this case, we only compare the relative efficiency of the neural networks, but we do not use the pre-computation of the grounded semantics which is mentioned in Section~\ref{section:technical-improvements}. Some instances required too much time to do so, and were thus excluded from the test set. Since all problems are not equivalently hard for a given AF, the exact test set is slightly different for each of them ($\DC$-$\complete$: 252 instances, $\DC$-$\stable$: 268 instances, $\DS$-$\preferred$: 217 instances and $\DS$-$\stable$: 259 instances).

Then, for each of these problems and AFs, we have executed the four different GNN models. Indeed, when running a GCN or a GAT, the output layer of the neural network does not only give an estimation of the acceptability for one specific argument, but for all the arguments in the AF, which means we can compare the precision of these four approaches for all the arguments.

A possible issue here, if we chose to directly present global results (for instance, providing the accuracy as $n/m$ with $n$ the number of correctly predicted arguments, and $m$ the total number of arguments over the full test set), would be related to the variability in the sizes of the AFs. For instance, if a neural network is particularly efficient for a given type of AFs which are generally large, it could ``hide'' the fact that the network does not perform very well on another type of AFs which are typically small. For this reason, we first compute the accuracy on a given instance passed to the GNN, where all its arguments are taken into account ($\theta_{\af} = n/m$ with $n$ the number of correctly predicted arguments in $\af$, and $m$ the number of arguments in $\af$). So we get an accuracy of the predicted acceptability on one given instance, and the end we compute the average accuracy across all AFs to get the global accuracy ($\sum_{\af} \theta_{\af} / N$ with $N$ the number of AFs in the test set).

\begin{table}[htb]
    \centering
    \begin{tabular}{ccccc}
       \toprule
       GNN & $\DC$-$\complete$ & $\DC$-$\stable$ & $\DS$-$\preferred$ & $\DS$-$\stable$ \\
       \midrule
       \afgcn v2 & $53$ ($48$;$76$) & $75$ ($57$;$76$) & $91$ ($20$;$\mathbf{99}$) & $64$ ($17$;$98$) \\
        \afgcn-P128 & $70.9$ ($38$;$95$) & $83$ ($65$;$89$) & $96$ ($52$;$\mathbf{99}$) & $71$ ($47$;$99$)  \\
        \afgcn-P11 & $67$ ($28$;$\mathbf{96}$) & $73$ ($42$;$90$) & $92$ ($26$;$97$) & $69$ ($31$;$99$)\\
        \afgcn-P128$^{-do}$ & $76$ ($52$;$89$) & $82$ ($61$;$89$) & $97$ ($63$;$\mathbf{99}$) & $72$ ($51$;$99$)\\
        \afgcn-P11$^{-do}$ & $74$ ($46$;$93$) & $83.0$ ($62$;$88$) & $96$ ($56$;$\mathbf{99}$) & $71$ ($47$;$98$)\\
        \afgat & $\mathbf{88}$ ($\mathbf{78}$;$90$) & $\mathbf{87}$ ($\mathbf{77}$;$\mathbf{92}$) & $\mathbf{98}$ ($\mathbf{80}$;$90$) & $\mathbf{73}$ ($\mathbf{59}$;$\mathbf{99}$)\\
        \bottomrule
    \end{tabular}
    \caption{Accuracy of the GNNs. In each cell, the first value corresponds to the global accuracy, and the number between parenthesis correspond respectively to the accuracy restricted to the positive and negative instances. Bold-faced values represent the  highest accuracy for a problem (and a type of instances).}
    \label{tab:results-accuracy-all-GNNs}
\end{table}

Our results are given in Table~\ref{tab:results-accuracy-all-GNNs}.
The main insight that we obtain from this experiment is the overall performance of \afgat, which obtains the best highest accuracy in most of cases.
We also observe that (except for $\DC$-$\complete$), our \afgcn-P128 performs better than \afgcn v2, which confirms the interest of our new features in the learning process. On the contrary, the version without randomized features (\afgcn-P11) is the least performing approach. The most plausible explanation is that dropout layers induce a major loss of information on a small number of features. This also explains why both approaches without dropout layers perform better than the other solvers based on GCNs.

\subsubsection{Comparison with the ICCMA 2023 Participants}
For the second experiment, we follow more closely the process of ICCMA 2023. In this case, for each AF, we use only one query argument for each AF (the same that was used in the competition). For this reason, there is no need to exclude AFs from the test set, since the ground truth is provided by ICCMA 2023 organizers.
This means that we can easily compare the results of an approach based on neural networks (here, \afgat) with the results of the best performers during the competition, namely \fargo{} and \harper. 
In this experiment, a timeout of $39$ seconds was enforced (instead of $60$ seconds at ICCMA 2023, because the experiments we conducted on a CPU $54\%$ higher clock rate than ICCMA 2023), and the pre-processing based on the grounded semantics is incorporated in all our GNNs based solver. The number of correctly solved instances are given in Table~\ref{tab:results-accuracy-iccma2023}. We observe that \fargo{} generally outperforms the other approaches, except for the skeptical acceptability under the stable semantics where our \afgat{} obtains the best results. However, the results are generally tight, and let us envision better results for future versions of \afgat{} with other parameters or another training dataset used in the learning process.

\begin{table}[htb]
    \centering
    \begin{tabular}{ccccc}
       \toprule
       GNN & $\DC$-$\complete$ & $\DC$-$\stable$ & $\DS$-$\preferred$ & $\DS$-$\stable$ \\
       \midrule
        \afgat & $265$ & $268$ & $297$ & $\mathbf{205}$ \\
        \harper & $220$ & $187$ & $300$ & $196$\\
        \fargo & $\mathbf{300}$ & $\mathbf{307}$ & $\mathbf{303}$ & $199$ \\
        \bottomrule
    \end{tabular}
    \caption{Number of correctly solved instances by the solvers on the ICCMA 2023 test set.}
    \label{tab:results-accuracy-iccma2023}
\end{table}

\section{Related Work}\label{section:related-work}
Besides \afgcn{}, other approaches have been proposed for approximate reasoning in abstract argumentation, including some based on machine learning.

As mentioned earlier, \cite{KuhlmannT19} proposed a first implementation of a GCN for solving the credulous acceptability problem under the preferred semantics. This preliminary work showed the potential interest of using GCNs for abstract argumentation, although the approach did not exhibit high enough performances for practical application (with accuracy scores remaining low for some types of instances). Then, \cite{CraandijkB20COMMA,CraandijkB20IJCAI} proposed a so-called Argumentation Graph Neural Network (AGNN) which approximates the acceptability of arguments under several semantics  with a high accuracy. The main difference between this approach and most other approaches (which participate to ICCMA competitions) is the size of the instances (less than $200$ arguments). Studying whether the AGNN  approach scales-up well with ICCMA instances (w.r.t. runtime and accuracy), and how it compares with our approaches, is an interesting task for future work.

Deep learning has also been used for other purposes, like extension enforcement \cite{DBLPCraandijkB22COMMA,CraandijkB22AAAI} or approximating gradual semantics of Bipolar AFs \cite{AnaissySSV24}. Even if these contributions are out of the scope of the present paper, a deeper analysis of these works could provide interesting insights for improving our algorithms.

Finally, we have already briefly mentioned another approach for approximate reasoning that participated to ICCMA 2023, namely \aripoter{} \cite{DelobelleMR23}. This solver combines the initial intuition of \harper{} (regarding the grounded extension) and the use of gradual semantics for evaluation arguments acceptability. In ICCMA 2023, it performed generally better than \afgcn{} but not as good as \harper{} and \fargo{}. An experimental comparison of our approaches based on neural networks with \aripoter{} is naturally an interesting future work.

\section{Conclusion}
We have shown how small changes of the GCN architecture from \afgcn{} can improve its performance on ICCMA 2023 benchmarks. Also, we have proposed a new approach based on GATs, which performs even better than all the GCN variants that we have tested, and is competitive with the best performers from ICCMA 2023. This opens interesting research questions regarding the used of neural networks in abstract argumentation.
A first natural extension of this work consists of empirically evaluating other variants of our GNN models ({\em e.g.} by testing other parameters for the behaviour of the dropout layers in the GCNs or the graph attentional layers in the GATs).
We plan to continue this line of work by taking into account other semantics ({\em e.g.} the ideal, semi-stable or stage semantics). We also wish to extend our approach to other kinds of abstract argumentation frameworks, like Incomplete AFs \cite{Mailly22} for which reasoning may be harder than standard AFs. Finally, it would be interesting to provide approximate reasoning methods for other problems ({\em e.g.} computing some extension of an AF).

 \section*{Acknowledgement}
 The second author received funding from the French National Research Agency (ANR grants ANR-22-CE23-0005 and ANR-22-CPJ1-0061-01). The authors thank Jérôme Delobelle for providing Example~\ref{example:gradual-semantics}.

%%%%%%%%%%% The bibliography starts:
\bibliographystyle{vancouver}
\bibliography{biblio}

\end{document}